\documentclass[10pt,twocolumn,letterpaper]{article}

\usepackage{wacv}              % Camera-ready version
\usepackage{booktabs}
\usepackage{colortbl}
\usepackage{multirow}
\usepackage{makecell}
\usepackage{pifont}
\definecolor{rcHeader}{RGB}{224,235,247}
\definecolor{rcStripe}{RGB}{245,247,250}
\definecolor{rcBlue}{RGB}{25,88,155}
\definecolor{rcRed}{RGB}{190,45,45}
\definecolor{rcGreen}{RGB}{20,120,82}
\newcommand{\rcbest}[1]{\textbf{\color{rcBlue}#1}}

\newcommand{\rcfail}[1]{\textcolor{rcRed}{#1}}

\newcommand{\cmark}{\ding{51}}

\newcommand{\rolecue}{\textsc{ROLECUE}}

\definecolor{wacvblue}{rgb}{0.21,0.49,0.74}
\usepackage[pagebackref,breaklinks,colorlinks,allcolors=wacvblue]{hyperref}

\def\wacvPaperID{2604} % *** Enter the WACV Paper ID here
\def\confName{WACV}
\def\confYear{2027}

\title{ROLECUE: Diagnosing Cross-Lingual Role-Binding Failures in Vision-Language Models}

\author{Niaz Mohaiman Abtahi \qquad Evan Maksud\\ Advanced AI Lab Limited\\ {\tt\small tahiniaz@gmail.com}\\[3pt]
{\small \href{https://github.com/NiazTahi/ROLECUE}{\raisebox{-0.15em}{\includegraphics[height=0.95em]{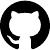}}\ GitHub} \quad
\href{https://huggingface.co/datasets/NiazTahi/ROLECUE-192}{\raisebox{-0.15em}{\includegraphics[height=0.95em]{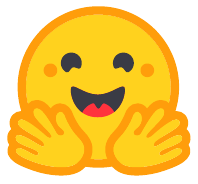}}\ Hugging Face}}}

\begin{document}
\maketitle
\begin{abstract}
Vision-language models (VLMs) can recognize events in images and parse sentences, but it remains unclear when they bind visual Agent--Patient structure to linguistic semantic roles across languages whose role cues depend on morphology, word order or construction.
We introduce \rolecue, a controlled diagnostic that pairs 192 validated two-person action images with German, Turkish, Japanese, English, Korean and Bangla sentences in canonical and patient-first orders.
The evaluation separates image-only role recognition, text-only role parsing, direct image--sentence matching, response-compliance behavior and an external modular composition of the model's own component predictions.
Across eight recent open-weight VLMs, \rolecue\ reveals a structured spectrum: some models fail visual or linguistic component tests, some largely solve the diagnostic, some collapse to constant semantic assent and several show component--joint gaps in which reliable image-only and text-only role predictions do not survive direct joint inference.
The largest non-degenerate gaps reach 44.9 percentage points, with final bootstrap intervals confirming that the effect is not a point-estimate artifact.
These results reveal multilingual component--joint performance gaps and acceptance biases hidden by aggregate VLM benchmarks and show why component diagnostics are necessary before interpreting joint multimodal errors.
\end{abstract}
    
\section{Introduction}
\label{sec:intro}

Vision-language models (VLMs) are increasingly evaluated as general-purpose systems that can answer questions about images, read text, and judge whether a caption matches a scene.
Yet many such evaluations do not isolate a basic semantic requirement: determining \emph{who did what to whom} across modalities.
This matters especially outside English.
Languages such as German, Turkish, Japanese, Korean, and Bangla can place the patient before the agent while grammatical marking or constructional cues still indicate the event roles.
For example, a Japanese sentence may begin with ``the blue-shirted man'' while still meaning that the red-shirted man pushes the blue-shirted man.
A model that treats the first mentioned participant as the agent can appear fluent while failing the semantic role binding needed for grounded understanding.

We study this failure mode with \rolecue, a controlled diagnostic for multilingual visual-linguistic role binding.
Each item contains a two-person action image with visually marked participants, one red and one blue, and a sentence whose grammatical role structure either matches or conflicts with the image.
The key design is diagnostic rather than benchmark-maximal: before interpreting a joint image--sentence error, we separately test whether the model can identify the visual agent from the image and whether it can identify the linguistic agent from the sentence alone.
Only then do we ask whether the sentence correctly describes the image.

\begin{figure*}[t]
  \centering
  \footnotesize
  \setlength{\tabcolsep}{3pt}
  \begin{tabular}{p{0.43\textwidth}p{0.43\textwidth}}
  \centering\textbf{Agent-first:} ``Red pushes blue.'' &
  \centering\textbf{Agent-first:} ``Red pushes blue.'' \tabularnewline
  \centering\includegraphics[width=0.31\textwidth]{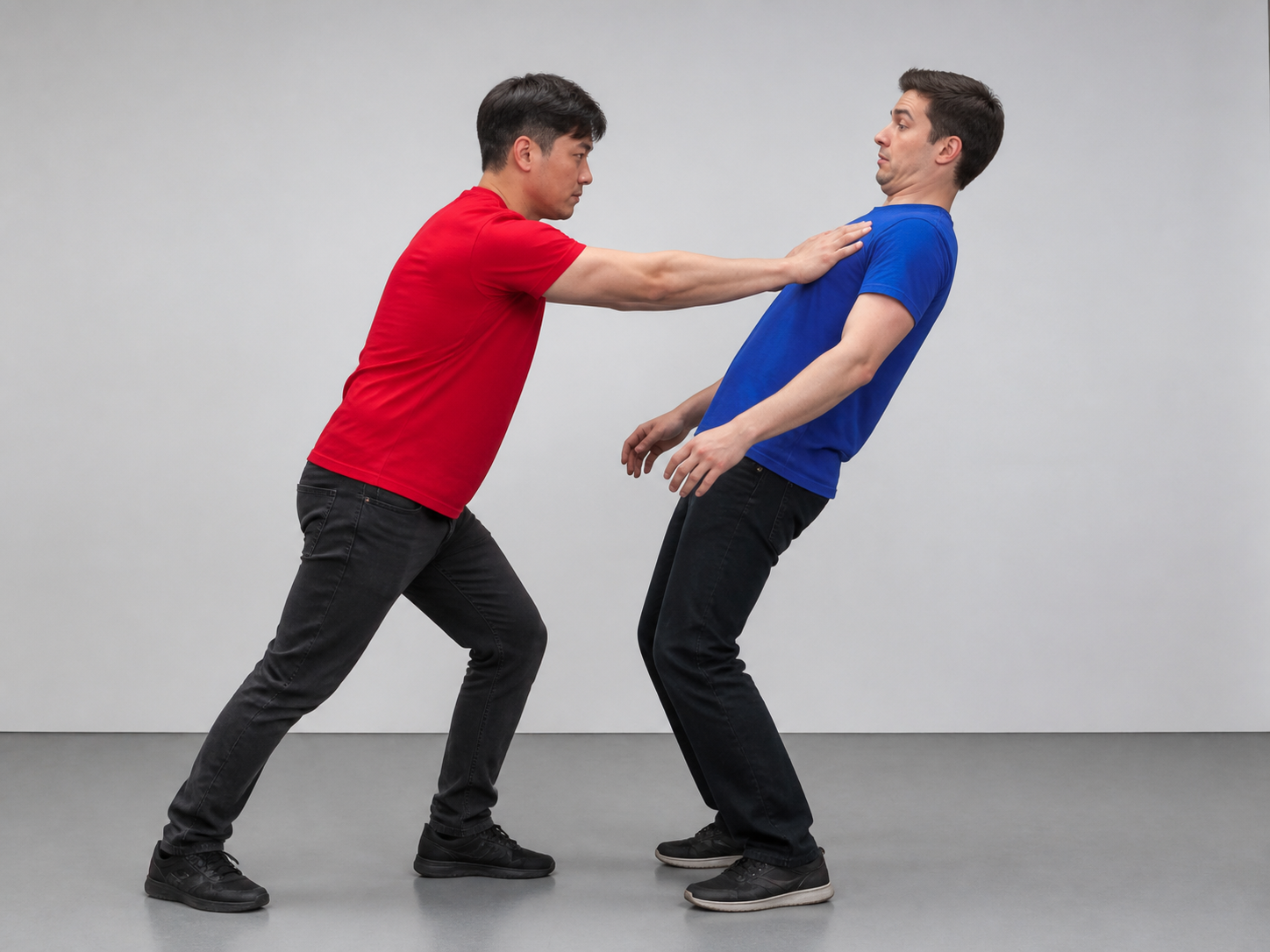} &
  \centering\includegraphics[width=0.31\textwidth]{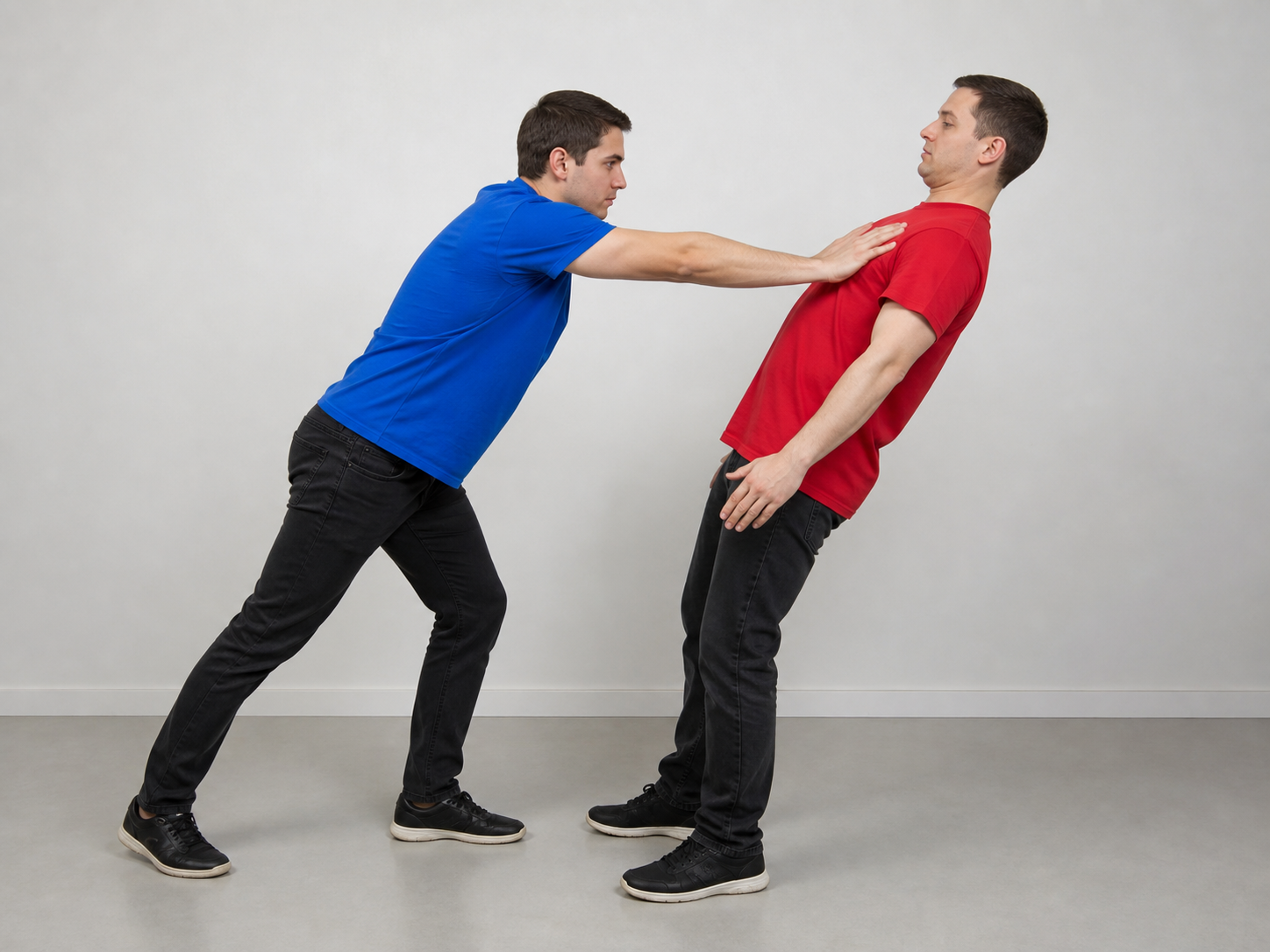} \tabularnewline
  \centering\textcolor{green!45!black}{\textbf{MATCH}}: visual agent is red &
  \centering\textcolor{red!70!black}{\textbf{MISMATCH}}: visual agent is blue \tabularnewline
  \addlinespace[1pt]
  \centering\textbf{Patient-first:} ``Blue is pushed by red.'' &
  \centering\textbf{Patient-first:} ``Blue is pushed by red.'' \tabularnewline
  \centering\includegraphics[width=0.31\textwidth]{figures/push_red_agent_left.png} &
  \centering\includegraphics[width=0.31\textwidth]{figures/push_blue_agent_left.png} \tabularnewline
  \centering\textcolor{green!45!black}{\textbf{MATCH}}: visual agent is red &
  \centering\textcolor{red!70!black}{\textbf{MISMATCH}}: visual agent is blue \tabularnewline
  \end{tabular}
  \caption{\rolecue\ independently varies visual role, grammatical role, and noun order. The figure shows the English condition for readability; the same controlled manipulation is instantiated with validated German, Turkish, Japanese, English, Korean, and Bangla sentences. Patient-first order makes first-noun heuristics fail while grammatical role cues still identify the sentence agent.}
  \label{fig:overview}
\end{figure*}

Our experiments reveal a structured pattern of errors.
Some models fail before the integration test: Aya-Vision-8B~\citep{dash2025ayavision}, for instance, does not pass the image-only role gate on several actions.
Other models pass visual gating, but fail to parse some patient-first sentences.
Another failure is response compliance: LLaVA-OneVision-7B reaches chance on several joint cells because it almost always accepts the sentence, so those cells are reported separately rather than counted as clean integration failures.
In selected cases, component questions are answered correctly while the direct joint image--sentence decision fails.
Qwen3-VL-8B~\citep{bai2025qwen3vl}, for example, reaches 99.0\% image-only accuracy on the full 192-image set, but drops to 63.3\% on Japanese agent-first and 65.1\% on Turkish patient-first joint matching despite perfect text-only scores in those settings.
At the same time, larger models substantially reduce the effect: Qwen3-VL-32B~\citep{bai2025qwen3vl} is nearly perfect across the same conditions, and InternVL3.5-38B~\citep{wang2025internvl35} passes the full visual gate and remains at or above 90\% in all patient-first joint settings.
The diagnostic reports component and joint behavior; it does not assume that all VLMs fail uniformly.

We further test, on targeted failure cells, whether the failures are merely prompt artifacts.
We compare the direct joint prompt against two mitigation prompts: one warning that word order may be noncanonical, and one asking the model to identify image and sentence agents before answering.
These prompts do not reliably help.
For Qwen3~\citep{bai2025qwen3vl} and Gemma~\citep{gemmateam2026gemma4}, they often reduce accuracy; for LLaVA~\citep{li2024llavaonevision} they largely preserve chance-level behavior; and for MiniCPM~\citep{yu2025minicpmv45} they can collapse performance far below its direct agent-first scores.
This supports a sharper diagnostic claim: asking for role-aware reasoning in one call is not equivalent to separately measuring the two role decisions and composing them outside the model.
When external composition succeeds but the direct joint answer fails, the component predictions support the correct comparison, though decision bias may also contribute.

Our contributions are:
\begin{itemize}
  \item We introduce \rolecue, a validated controlled diagnostic for cross-lingual visual-linguistic Agent--Patient role binding.
  \item We separate visual recognition, text-only role parsing, and joint multimodal role matching, preventing visual failures from being mistaken for integration failures.
  \item We introduce an external modular comparison that presents the component--joint gap directly and distinguishes missing component knowledge from failed composition of available role decisions.
  \item We evaluate eight recent open-weight VLMs as main models, spanning Qwen, InternVL, Gemma, LLaVA, and MiniCPM families, and report additional diagnostics for Aya, GLM, and Phi.
  \item We show that failures are not uniform: large recent models can largely solve the benchmark, while smaller and older models expose sharp language- and order-specific bottlenecks.
\end{itemize}

\section{Related Work}
\label{sec:related}

\paragraph{Vision-language models.}
Modern VLMs build on contrastive image--text pretraining~\citep{radford2021clip}, instruction-tuned multimodal assistants such as LLaVA~\citep{liu2023llava}, and increasingly capable open model families including Qwen-VL~\citep{bai2023qwen,bai2025qwen25vl,bai2025qwen3vl}, InternVL~\citep{chen2024internvl,wang2025internvl35}, MiniCPM-V~\citep{yu2025minicpmv45}, LLaVA-OneVision~\citep{li2024llavaonevision}, Gemma~\citep{gemmateam2026gemma4}, Aya~\citep{dash2025ayavision}, GLM~\citep{hong2025glm41v}, and Phi~\citep{aneja2026phi4rv}.
These systems are often evaluated with broad visual question answering and visual reasoning tasks~\citep{antol2015vqa,hudson2019gqa,suhr2019nlvr2}, as well as captioning, OCR, and chart/document understanding tasks.
\rolecue\ instead targets a narrow semantic operation: binding visual event roles to linguistic roles under cross-lingual word-order variation.

\paragraph{Compositional and relational evaluation.}
Prior work has shown that vision-language models can struggle with compositionality, relation understanding, and controlled caption--image contrasts~\citep{hendricks2021decoupling,lewis2024clipbind}.
Winoground~\citep{thrush2022winoground} counterbalances lexical content while reversing relations, VALSE~\citep{parcalabescu2022valse} constructs linguistic foils, ARO~\citep{yuksekgonul2023aro} and SugarCrepe~\citep{hsieh2023sugarcrepe} target order/relation sensitivity and benchmark artifacts, and CREPE~\citep{ma2023crepe} measures systematic and productive composition.
These benchmarks establish that aggregate retrieval or VQA performance~\citep{antol2015vqa} does not guarantee structure-sensitive grounding.
\rolecue\ shares their controlled spirit but asks a different question: whether an explicit visual Agent--Patient assignment, in the predicate-argument sense used by semantic-role resources such as PropBank~\citep{palmer2005propbank}, can be bound to a morphosyntactically determined sentence-role assignment.
The distinction is important because the relevant linguistic cue is not merely English word order.
In the patient-first condition, both captions mention the same participants and action, and only grammatical role cues determine which participant is the agent.

\paragraph{Multilingual grounding.}
Cross-lingual VQA resources such as xGQA, MaXM, and CVQA broaden visual evaluation beyond English~\citep{pfeiffer2022xgqa,changpinyo2023maxm,romero2024cvqa}, while EXAMS-V tests multilingual multimodal reasoning over naturally occurring educational material~\citep{das2024examsv}.
Their breadth is complementary to our controlled design.
Different questions, images, topics, and cultural contexts can vary together across languages, making a language-level accuracy difference difficult to localize.
Recent multilingual binding diagnostics similarly point to fragile cross-lingual association in VLMs~\citep{chakraborty2026m2bind}.
\rolecue\ instead holds the visual event and intended proposition fixed while changing language and argument order.
It uses six languages---German, Turkish, Japanese, English, Korean, and Bangla---that realize the role contrast through language-appropriate case, postpositional, possessive, or passive-construction cues.
The goal is not to rank languages or estimate general multilingual ability; it is to expose whether VLMs integrate visual roles with overt grammatical evidence when a first-noun shortcut is invalid.

\paragraph{Diagnostic decomposition.}
Most multimodal benchmarks score only the final joint answer.
That design cannot distinguish a model that misrecognizes the image, one that misparses the sentence, and one that solves both components but combines them incorrectly.
\rolecue\ treats image-only and text-only performance as measured preconditions for interpreting the joint score.
The resulting decomposition is analogous to a controlled intervention: visual content, sentence-role structure, and their compatibility are varied independently, while answer labels are counterbalanced.
Strong component scores identify a component--joint gap, whose causes may include decision bias and prompt sensitivity.

\section{The \rolecue\ Diagnostic}
\label{sec:method}

\subsection{Stimuli}

\rolecue\ contains 192 generated two-person action images.
Each image depicts a red-shirted and a blue-shirted person involved in one transitive action.
We use 12 action types: bandaging, carrying, chasing, dragging, feeding, hugging, kicking, lifting, photographing, pushing, tickling, and washing.
For each action we vary the visual agent color and side, producing role-reversal pairs.
Human validators judged the visual role in all 192 images; all images passed the human visual validation protocol.
The visual set is compact by design, but its controlled crossings produce a substantially larger evaluation.
For models run on the full set, the 192 validated images yield 4,608 collapsed joint image--sentence decisions per model after crossing language, word order, match status, sentence agent color, and answer-map counterbalancing.
For no-chasing analyses, the corresponding count is 4,224 collapsed joint decisions per model.
The text-only probe contains 24 distinct sentences per language--order cell for full-set models and 22 for no-chasing models, repeated for alignment with the joint trials.
Repeated text-only manifest rows are retained only to align bookkeeping with the corresponding joint items.

\subsection{Controlled Construction and Human Validation}

The 192-image pool comprises two independently generated waves of 96 images rather than an original set plus horizontal flips.
Within every action, the design balances agent color (red/blue), agent side (left/right), and four visual seeds.
This crossing prevents color, position, or a single rendering from determining the role label.
The visual seed identifiers (\texttt{s1}--\texttt{s4}) are frozen stimulus IDs stored in every manifest row together with action, role labels, image path, language condition, answer map, and prompt.
The original 96 images received three recorded visual assessments with unanimous intended-agent judgments. For the 96 new images, the two archived submissions are identical and contain the same annotator ID; the available records therefore do not establish two independent assessments. All retained images have at least one recorded intended-agent judgment.

Language validation is likewise performed before model inference.
The language-review form instructs participants to assess their native language; it shows the sentence and intended roles and asks about grammaticality, meaning, and naturalness on a five-point scale, with optional corrections. The archived results contain two validator IDs per language; recruitment and independence could not be established from the forms alone.
All 12 redesigned Korean and Bangla templates in the final repair round passed two-validator checks for grammar and intended meaning.
We retain language-appropriate marking rather than forcing every language into one case frame: for example, Korean bandaging uses a nominative--dative contrast and Bangla photographing uses a nominative--genitive/possessor contrast.
Consequently, the shared experimental variable is 
\emph{overtly recoverable semantic role}, not a claim that all six languages instantiate identical morphology.

\subsection{Languages and Role Manipulations}

For each visual item, we construct German, Turkish, Japanese, English, Korean, and Bangla sentences.
Each sentence specifies the same action and participant colors, while varying whether the agent appears first (agent-first) or the patient appears first (patient-first).
The patient-first condition is the critical test: the first noun is not the agent, so a first-noun heuristic predicts the wrong role.
The language set is deliberately heterogeneous: English provides a passive baseline, German/Turkish/Japanese test case or postpositional cues, and Korean/Bangla add typologically and script-diverse validated constructions.

\subsection{Inference Tasks}

We evaluate three tasks.
The image-only task asks which person performs the action in the image.
The text-only task asks which person performs the action according to the sentence.
The joint task asks whether the sentence correctly describes which person performs the action in the image.
All tasks use two answer options and reverse the A/B mapping.
For example:
\begin{quote}\small
Which person is pushing the other person? A: The person in red. B: The person in blue. Answer with only A or B.
\end{quote}
and
\begin{quote}\small
Does the sentence correctly describe which person is performing the action on the other person in the image? A: Yes. B: No. Answer with only A or B.\\
Sentence:\\
\vspace{1em}
\includegraphics[width=\linewidth]{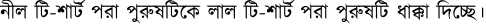}
\end{quote}
The reversed mappings allow us to collapse results by semantic margin rather than raw label preference.

\subsection{Scoring and Statistical Unit}

Each semantic item is queried twice with the two answer mappings reversed.
For models exposing first-token scores, let $m_1$ and $m_2$ be the A/B logit differences, each oriented toward the same semantic choice.
We score the item by the antisymmetric margin $(m_1+m_2)/2$ after orienting both margins toward the same semantic decision.
For interfaces without comparable logits, the two parsed hard answers must agree semantically; conflicting or unparsed answers are counted as incorrect. With logits, generated-answer compliance is reported separately: the primary score can be computed even when the generated text is not parsed. Exact zero margins select red for text and match for joint; the image-gate code selects blue on a tie.

Accuracy is reported over collapsed semantic items, never over the duplicated answer mappings.
For uncertainty on modular gains and component--joint gaps, we resample visual image keys within each model--language--order condition so that all matched sentences and answer mappings derived from the same image remain in one bootstrap cluster.
This avoids treating counterbalanced prompts as independent observations.
Text-only rows are repeated to align with image-based trials; when predictions vary, their aggregate reflects these repeated trials rather than independent sentence templates.
The primary tables report all balanced cells rather than significance-filtered subsets; the component-competent modular view applies the predeclared 85\% image-only and text-only thresholds solely to identify settings in which an integration interpretation is warranted.

\subsection{Enforced Modular Inference}

The direct joint task asks the model to combine the image and sentence in one multimodal call.
To present whether a joint error reflects missing component knowledge or failed composition, we also compute an external modular prediction.
For each joint item, we take the model's collapsed image-only prediction, take its collapsed text-only prediction for the corresponding sentence, and compare the two predicted agent colors outside the model.
The modular prediction is \emph{match} if the predicted visual agent color equals the predicted sentence agent color, and \emph{mismatch} otherwise.
If both component predictions are correct but the direct joint answer is wrong, the item is counted as a component--joint disagreement.
The modular score is a deterministic composition of the model's own component predictions, not a separate VLM call; its purpose is to make the component--joint gap explicit.
For modular gains, we report percentile bootstrap confidence intervals over visual image keys within each model-language-order group.

\subsection{Visual Gate}

Static chasing scenes can be visually ambiguous. We therefore first apply an image-only gate and use a no-chasing subset for models that fail this action.
This gate is computed before inspecting joint image--sentence results and uses only the image-only role question.
Actions that a model cannot solve reliably are excluded from that model's primary integration table and reported separately.
This prevents visually ambiguous actions from being misinterpreted as cross-modal integration failures.

\section{Experiments}
\label{sec:experiments}

\subsection{Models}

The main experiments evaluate eight clean open-weight VLMs: Qwen3-VL-8B, Qwen3-VL-32B, Qwen2.5-VL-32B, InternVL3.5-8B, InternVL3.5-38B, LLaVA-OneVision-7B, MiniCPM-V-4.5, and Gemma-4-12B-it.
This set gives both cross-family diversity and within-family scale comparisons.
We also report diagnostic attempts with Aya-Vision-8B~\citep{dash2025ayavision}, GLM-4.1V-9B-Thinking~\citep{hong2025glm41v}, and Phi-4-reasoning-vision-15B~\citep{aneja2026phi4rv} in the supplement.
Aya fails the visual precondition, GLM produces analyzable first-token logits but not strict A/B hard responses, and Phi was excluded because the available runtime and remote-code stack were incompatible.
All clean main models are run in bfloat16 without quantization.
For models that expose first-token A/B scores, we collapse reversed answer mappings using antisymmetric semantic margins.
For MiniCPM, whose chat API does not expose comparable first-token logits, we collapse by hard A/B predictions.

\subsection{Prompting Mitigations}

To test whether failures are prompt artifacts, we evaluate two joint-task mitigation prompts.
The \emph{conflict-warning} prompt explicitly warns that the sentence may use noncanonical word order and instructs the model to use grammatical case marking.
The \emph{role-first} prompt asks the model to identify the image agent, identify the sentence agent, and compare them before answering.
Both prompts still require a single A/B output.

\section{Results}
\label{sec:results}

\subsection{Visual Role Gate}

\begin{table*}[t]
\centering
\small
\setlength{\tabcolsep}{3.5pt}
\begin{tabular*}{\textwidth}{@{\extracolsep{\fill}}lccrrrr@{}}
\toprule
\rowcolor{rcHeader}
\textbf{Model} & \textbf{Primary set} & \textbf{Images} & \textbf{Primary acc.} & \textbf{Full acc.} & \textbf{Chasing} & \textbf{Gate} \\
\midrule
\rowcolor{rcStripe} Qwen3-VL-8B & full & 192 & \rcbest{99.0} & \rcbest{99.0} & 87.5 & \textcolor{rcGreen}{\cmark} \\
Gemma-4-12B-it & full & 192 & \rcbest{100.0} & \rcbest{100.0} & \rcbest{100.0} & \textcolor{rcGreen}{\cmark} \\
\rowcolor{rcStripe} InternVL3.5-8B & no-chasing & 176 & \rcbest{99.4} & 91.7 & \rcfail{6.2} & \textcolor{rcRed}{no-chasing} \\
LLaVA-OneVision-7B & no-chasing & 176 & \rcbest{100.0} & 96.4 & 56.2 & \textcolor{rcRed}{no-chasing} \\
\rowcolor{rcStripe} MiniCPM-V-4.5 & no-chasing & 176 & \rcbest{100.0} & 96.4 & 56.2 & \textcolor{rcRed}{no-chasing} \\
Qwen2.5-VL-32B & no-chasing & 176 & \rcbest{99.4} & 97.9 & 81.2 & \textcolor{rcRed}{no-chasing} \\
\rowcolor{rcStripe} Qwen3-VL-32B & full & 192 & \rcbest{99.5} & \rcbest{99.5} & 93.8 & \textcolor{rcGreen}{\cmark} \\
InternVL3.5-38B & full & 192 & \rcbest{100.0} & \rcbest{100.0} & \rcbest{100.0} & \textcolor{rcGreen}{\cmark} \\
\bottomrule
\end{tabular*}
\caption{Image-only visual-role gate on \rolecue-192. Primary acc. is computed on the set used for integration analysis: all 192 images for full-set models and the 176-image no-chasing subset for restricted models. Full acc. is shown for transparency, and Chasing gives the action-level score that determines the subset.}
\label{tab:visual-gate}
\end{table*}

Table~\ref{tab:visual-gate} shows that visual competence cannot be assumed, but also that the final clean main set is visually strong after the predeclared gate.
The action \emph{chasing} is particularly fragile for several models, likely because a static frame makes the chaser/chased relation less explicit.
Because this failure is detected from image-only predictions before any joint analysis, the no-chasing split is a visual-control decision rather than a post-hoc selection based on joint performance.
We therefore use the full set for Qwen3-VL-8B, Gemma-4-12B-it, Qwen3-VL-32B, and InternVL3.5-38B, and the no-chasing set for InternVL3.5-8B, LLaVA-OneVision-7B, MiniCPM-V-4.5, and Qwen2.5-VL-32B.
Aya is not silently discarded: it is reported as a failed-gate diagnostic model, showing that some VLMs do not meet the visual precondition needed for an integration claim.
GLM is likewise reported as a protocol diagnostic because its thinking-style outputs do not satisfy the strict direct A/B hard-response protocol.

\subsection{Joint Role Binding}

\begin{figure*}[t]
    \centering
    \includegraphics[width=\textwidth]{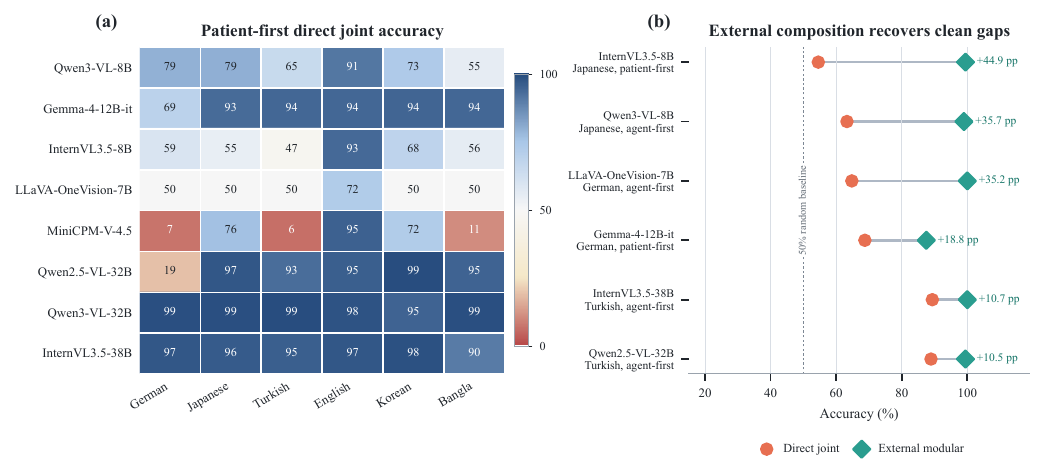}
    \caption{\rolecue-192 diagnostic overview. (a) Patient-first direct-joint accuracy across eight open-weight models and six language--construction conditions; the full or no-chasing evaluation set follows the independently determined image-only gate. (b) Direct joint inference versus external modular composition in non-degenerate component-competent settings, where image-only and text-only accuracy are at least 85\% and the direct joint output is not a constant semantic response. We retain each model's strongest qualifying recovery and show the six largest model-level gains. The dashed line is the 50\% random baseline; gains are percentage points. Complete agent-first, patient-first, text-only, response-compliance, and modular results appear in the supplement.}
    \label{fig:diagnostic-overview}
\end{figure*}

Figure~\ref{fig:diagnostic-overview}(a) gives the main result in compact form; the supplement additionally reports agent-first and patient-first accuracy for all models on the common no-chasing set.
Performance varies widely across model families and scales.
The larger Qwen3-VL-32B and InternVL3.5-38B models are strong across nearly all conditions, demonstrating that \rolecue\ is solvable by current open-weight systems.
At the same time, the figure exposes failures that aggregate benchmarks would hide.
Qwen2.5-VL-32B reaches 97.9\% image-only accuracy, yet collapses to 18.8\% on German patient-first joint matching on the visual-control subset.
MiniCPM is strong on many agent-first conditions, but collapses to 6.8\%, 6.0\%, and 10.8\% on German, Turkish, and Bangla patient-first joint matching.
Qwen3-VL-8B and InternVL3.5-8B show additional language-specific asymmetries, while LLaVA remains close to chance in most no-chasing joint conditions.
These patterns are consistent with model, language, and word-order-specific role-binding weaknesses rather than a single uniform failure mode.

\paragraph{Agent-first is a necessary control, not an easier benchmark split.}
Patient-first results are central because they break the first-noun heuristic, but their interpretation depends on the paired agent-first condition.
If a model fails both orders, the problem may be action recognition, general sentence understanding, or response compliance.
If it succeeds only when the agent is mentioned first, the contrast localizes an order-sensitive shortcut.
The supplement therefore reports the full agent-first matrix rather than hiding it behind the patient-first summary.
English serves as an anchor condition, but not always as the easiest one: Qwen3-VL-8B, for example, performs poorly on English patient-first text-only passives while doing much better on the corresponding joint prompt.
This shows that component probes are diagnostic measurements, not lower bounds on joint performance.
We do not average the six languages into a single headline score because that would obscure exactly the structured differences the diagnostic is designed to reveal.

\begin{table}[t]
\centering
\scriptsize
\setlength{\tabcolsep}{2.4pt}
\begin{tabular*}{\columnwidth}{@{\extracolsep{\fill}}lrrrrrr@{}}
\toprule
\rowcolor{rcHeader}
\textbf{Model} & \textbf{German} & \textbf{Japanese} & \textbf{Turkish} & \textbf{English} & \textbf{Korean} & \textbf{Bangla} \\
\midrule
\rowcolor{rcStripe} Qwen3-VL-8B & 87.5 & \rcbest{100.0} & \rcbest{100.0} & \rcfail{16.7} & \rcbest{100.0} & \rcfail{45.8} \\
Gemma-4-12B-it & 87.5 & 70.8 & \rcbest{100.0} & 87.5 & \rcbest{100.0} & \rcbest{100.0} \\
\rowcolor{rcStripe} InternVL3.5-8B & 59.1 & \rcbest{100.0} & 54.5 & \rcfail{36.4} & 95.5 & \rcfail{45.5} \\
LLaVA-OneVision-7B & \rcfail{45.5} & 95.5 & \rcfail{36.4} & 77.3 & 90.9 & 54.5 \\
\rowcolor{rcStripe} MiniCPM-V-4.5 & \rcfail{15.6} & \rcfail{41.8} & \rcfail{1.4} & \rcfail{11.4} & \rcfail{41.5} & \rcfail{12.5} \\
Qwen2.5-VL-32B & \rcfail{36.4} & \rcbest{100.0} & \rcbest{100.0} & \rcbest{100.0} & \rcbest{100.0} & \rcbest{100.0} \\
\rowcolor{rcStripe} Qwen3-VL-32B & \rcbest{100.0} & \rcbest{100.0} & \rcbest{100.0} & \rcbest{100.0} & \rcbest{100.0} & \rcbest{100.0} \\
InternVL3.5-38B & \rcbest{100.0} & \rcbest{100.0} & \rcbest{100.0} & \rcbest{100.0} & \rcbest{100.0} & \rcbest{100.0} \\
\bottomrule
\end{tabular*}
\caption{Patient-first text-only role-parsing accuracy. Values are computed over unique sentence templates per language/order condition (24 for full-set models, 22 for no-chasing models). Comparing this table with Figure~\ref{fig:diagnostic-overview}(a) separates sentence-role parsing failures from joint matching failures.}
\label{tab:text-only}
\end{table}

\begin{table*}[t]
\centering
\small
\setlength{\tabcolsep}{3.0pt}
\begin{tabular*}{\textwidth}{@{\extracolsep{\fill}}lllrrrrr@{}}
\toprule
\rowcolor{rcHeader}
\textbf{Model} & \textbf{Language} & \textbf{Order} & \textbf{Image} & \textbf{Text} & \textbf{Joint} & \textbf{Pred. match} & \textbf{Gap} \\
\midrule
\rowcolor{rcStripe} InternVL3.5-8B & Japanese & Patient-first & \rcbest{99.4} & \rcbest{100.0} & 54.5 & 94.9 & \rcfail{\textbf{44.9}} \\
InternVL3.5-8B & Turkish & Agent-first & \rcbest{99.4} & \rcbest{100.0} & 56.2 & 90.9 & \rcfail{\textbf{43.2}} \\
\rowcolor{rcStripe} InternVL3.5-8B & Bangla & Agent-first & \rcbest{99.4} & \rcbest{100.0} & 56.5 & 93.5 & \rcfail{\textbf{42.9}} \\
Qwen3-VL-8B & Japanese & Agent-first & \rcbest{99.0} & \rcbest{100.0} & 63.3 & 84.1 & \rcfail{\textbf{35.7}} \\
\rowcolor{rcStripe} LLaVA-OneVision-7B & German & Agent-first & \rcbest{100.0} & \rcbest{100.0} & 64.8 & 85.2 & \rcfail{\textbf{35.2}} \\
Qwen3-VL-8B & Turkish & Patient-first & \rcbest{99.0} & \rcbest{100.0} & 65.1 & 73.4 & \rcfail{\textbf{33.9}} \\
\rowcolor{rcStripe} Qwen3-VL-8B & Turkish & Agent-first & \rcbest{99.0} & \rcbest{100.0} & 71.1 & 74.7 & \rcfail{\textbf{27.9}} \\
InternVL3.5-8B & Korean & Patient-first & \rcbest{99.4} & 95.5 & 68.2 & 79.0 & \rcfail{\textbf{26.7}} \\
\rowcolor{rcStripe} Qwen3-VL-8B & Korean & Patient-first & \rcbest{99.0} & \rcbest{100.0} & 72.7 & 72.1 & \rcfail{\textbf{26.3}} \\
Qwen3-VL-8B & Bangla & Agent-first & \rcbest{99.0} & \rcbest{100.0} & 77.3 & 70.6 & \rcfail{\textbf{21.6}} \\
\bottomrule
\end{tabular*}
\caption{Largest non-degenerate integration gaps among settings with strong component performance. Gap is modular accuracy minus direct joint accuracy. Pred. match is the percentage of direct-joint trials where the model predicts that the sentence matches the image; near-constant match cells are reported as response-compliance failures in the supplement.}
\label{tab:fusion-gaps}
\end{table*}

Text-only results (Table~\ref{tab:text-only}) and the integration-gap summary (Table~\ref{tab:fusion-gaps}) show that failures have different sources across models.
Some models, such as MiniCPM and Qwen2.5-VL-32B in German patient-first, already struggle to parse the sentence role structure.
Others show high text-only competence but degrade in joint matching; response biases may contribute to this gap.
The gap table also reports the direct-joint predicted-match rate.

The gap table reports externally composed modular accuracy minus direct joint accuracy. This is a behavioral comparison of measured outputs; semantic-match bias may contribute to the gap.
For instance, InternVL3.5-8B drops to 54.5\% on Japanese patient-first despite 99.4\% image-only and 100.0\% text-only accuracy on the same no-chasing subset.
Conversely, MiniCPM's severe patient-first joint failures are not automatically labeled integration failures because its text-only accuracy is already below chance in several of the same cells.
LLaVA's constant-assent cells are excluded from this headline gap table and reported in the supplement as response-compliance failures.

On the common 176-image no-chasing set, InternVL3.5-8B accepts 99.4\% of Japanese patient-first matches and 90.3\% of mismatches. In an exploratory leave-one-action-out test, selecting a threshold on the other ten actions raises accuracy from 54.5\% to 87.8\%; this calibrated score uses labeled development data and is not directly comparable to the zero-shot score. The result indicates that decision bias contributes to this gap, without identifying a unique failure mechanism.

\subsection{External Modularity Presents the Component--Joint Gap}

Figure~\ref{fig:diagnostic-overview}(b) visualizes the component--joint gap in non-degenerate, component-competent settings; the corresponding numerical table with bootstrap confidence intervals is reported in the supplement.
In several settings, the model produces correct component predictions.
However, the same model fails when asked to answer the joint image--sentence question directly.
External modular comparison composes those same component predictions, often recovering the correct decision.
Because this modular score is computed from image-only and text-only outputs, it should be read as a transparent presentation of the component--joint gap rather than as an independent intervention.
The modular analysis is also informative when it does \emph{not} improve accuracy.
Qwen2.5-VL-32B's German patient-first row illustrates a component limitation: modular inference cannot recover what the text-only role parser does not supply.
Conversely, InternVL3.5-38B shows that even a strong large model can retain a small recoverable integration gap, while Qwen3-VL-32B leaves little room for modular improvement.

\subsection{Prompting Does Not Reliably Fix Role Binding}

The supplement reports targeted mitigation runs for a natural objection: perhaps the direct prompt is too terse.
The two tested instruction variants do not provide consistent improvements; few-shot prompting was not evaluated.
These prompts are not intended to exhaust the space of possible prompt engineering; rather, they test whether two natural instruction-level repairs are sufficient.
Across targeted failure settings, conflict warnings provide only small and inconsistent changes, and one-call role-first prompting often reduces accuracy toward chance.
For instance, Qwen3-VL-8B falls on Japanese agent-first items from 62.5\% direct accuracy to 55.2\% with conflict warning and 54.2\% with role-first prompting; its Turkish patient-first cell improves only slightly under warning, 62.5\% to 65.1\%, but drops to 50.5\% under role-first prompting.
InternVL3.5-8B shows the same asymmetry, while LLaVA-OneVision-7B remains at chance in selected cells, consistent with response-compliance failure rather than prompt-recoverable binding.
This supports the distinction between \emph{asking} a model to decompose the problem and \emph{enforcing} the decomposition outside the model.

\subsection{Scale Helps, but Does Not Define the Effect}

Within the evaluated families, Qwen3-VL-32B and InternVL3.5-38B improve over their 8B counterparts; these comparisons do not isolate parameter count from other model differences.
Cross-family and cross-generation comparisons also reflect differences in architecture and training: Qwen2.5-VL-32B retains a severe German patient-first failure, whereas Gemma-4-12B-it is substantially stronger in that cell.
The conclusion is behavioral rather than architectural: stronger systems often reduce the bottleneck, but component tests remain necessary to establish whether a particular model binds roles reliably.
Near-ceiling models also act as positive controls, demonstrating that the stimuli and scoring rule are solvable rather than intrinsically ambiguous.

\section{Discussion}
\label{sec:discussion}

\rolecue\ separates three phenomena that are often conflated.
First, a model may fail visual role recognition, as Aya does on the visual gate.
Second, a model may fail sentence-role parsing, as MiniCPM does on several patient-first text-only settings.
Third, a model may comply poorly with the direct joint format, as LLaVA's constant semantic assent shows in several languages.
Fourth, a model may possess visual and linguistic competence but fail the joint comparison.
The fourth case is a behavioral component--joint gap, potentially affected by several mechanisms.
Operationally, the diagnostic measures visual and linguistic components, response compliance, semantic-match bias, and component--joint gaps; these observations do not uniquely identify the cause of an error.
The modular analysis makes this distinction concrete.
External composition shows that the component probes can support a correct comparison, but it does not isolate a unique internal cause of the direct error; response and decision-threshold biases can also contribute.

The results also show why \rolecue\ should be interpreted as a diagnostic rather than a leaderboard.
Large recent models reduce many failures: Qwen3-VL-32B is nearly perfect, and InternVL3.5-38B remains strong across all languages and orders.
But scale does not collapse the diagnostic into a trivial benchmark.
Qwen2.5-VL-32B exposes a severe German patient-first weakness, InternVL3.5-38B still has small residual gaps, and several smaller models show large order-conditioned failures.
These patterns suggest that role binding should be evaluated explicitly rather than assumed from broad VQA or caption-matching performance.
As an additional closed-model reference, the supplement reports OpenAI GPT-5.6-Luna, GPT-5.6-Terra, and GPT-5.6-Sol under the same hard A/B protocol.
The reference results show generally high scores with variation across language conditions; coverage differs across languages as noted in the supplement.
This reinforces the intended use of \rolecue: not as a claim that every model fails, but as a test that localizes where role-binding behavior succeeds or breaks.

\subsection{What the Diagnostic Establishes}

Role reversal preserves participants and action vocabulary; answer-map reversal separates semantics from output-token preference; and component measurement checks vision and language before interpreting a joint error.
The strongest evidence is therefore not low joint accuracy alone, but high image-only and text-only accuracy, low non-degenerate direct-joint accuracy, and modular recovery.
The observed component decisions may be available even when a direct answer overweights a plausible match or the first-mentioned entity; these probes do not establish the model's internal representations.
Because modular comparison uses the model's own component predictions, recovery injects no new recognizer or parser; it changes only where the role assignments are combined, and should be interpreted as an external diagnostic composition.

\subsection{Reproducibility and Intended Use}

The release is organized around frozen manifests, not only aggregate tables.
The \href{https://huggingface.co/datasets/NiazTahi/ROLECUE-192}{dataset} provides the images and frozen manifests; the \href{https://github.com/NiazTahi/ROLECUE}{code repository} provides prompts, analysis scripts, and archived model outputs.
The full manifest contains 384 image-only prompt rows and 9,216 joint rows before answer-map collapse; text-only rows are aligned to those joint items but reduce to 288 unique validated sentences.
Model-specific inference settings and output formats affect scoring; bootstrap summaries use fixed analysis seeds.
Model-specific preprocessing and chat-template details are retained because they are part of the evaluated system and can affect strict-answer behavior.

\rolecue\ is intended for diagnosis and regression testing, not for ranking languages or claiming coverage of unrestricted events.
Human validation establishes that roles are recoverable in this stimulus set; it does not make the generated, color-coded scenes representative of people, cultures, or natural photography.

\section{Limitations}
\label{sec:limitations}

\rolecue\ is a controlled diagnostic with 192 generated images, 12 actions, and six languages; its color-marked, male-presenting participants limit demographic and visual diversity.
Human role-clarity validation does not establish generalization to natural scenes or broader populations, and repeated conditions do not add independent images.
Instructions are in English for every language; mixed-language prompting may affect results, and target-language instructions were not tested.
Static \emph{chasing} scenes can be ambiguous and are excluded from primary integration analyses only for models that fail the image-only gate on that action.
Model-specific chat templates and reasoning modes can also complicate strict A/B evaluation, as seen with GLM-4.1V-9B-Thinking and Phi-4-reasoning-vision-15B.

\section{Conclusion}
\label{sec:conclusion}

We introduced \rolecue, a controlled diagnostic for cross-lingual visual-linguistic role binding.
By separating image-only, text-only, and joint inference, \rolecue\ reveals failures that are hidden by aggregate multilingual VLM benchmarks.
Across recent open VLMs, strong visual recognition and sentence understanding do not always compose into reliable multimodal role binding.
Moreover, explicit role-aware prompting does not reliably fix the failures.
These findings motivate more precise role-binding evaluations.

% This section is intentionally left empty in the ROLECUE draft.

% This section is intentionally left empty in the ROLECUE draft.

\clearpage
{
    \small
    \bibliographystyle{ieeenat_fullname}
    \bibliography{main}
}

\end{document}